\documentclass[letterpaper, 10 pt, conference]{ieeeconf}  

\IEEEoverridecommandlockouts                              

\usepackage[T1]{fontenc}

\usepackage[table]{xcolor}
\usepackage{graphics} 
\usepackage{epsfig} 
\usepackage{mathptmx} 
\usepackage{times} 
\usepackage{amsmath} 
\usepackage{amssymb}  
\usepackage{subcaption}
\usepackage{hyperref}
\usepackage{graphicx}
\usepackage{bm}
\usepackage{gensymb}
\usepackage{todonotes}
\usepackage{multirow}

\usepackage[noadjust]{cite}

\newcolumntype{C}[1]{>{\centering\let\newline\\\arraybackslash}m{#1}}

\usepackage{fancyhdr}
\fancypagestyle{specialfooter}{%
  \fancyhf{}
  
  \fancyfoot[L]{\footnotesize This work was accepted by the IEEE/RSJ International Conference on Intelligent Robots and Systems, Abu Dhabi, UAE, 2024.\linebreak
  © 2024 IEEE. Personal use of this material is permitted.  Permission from IEEE must be obtained for all other uses, in any current or future media, including reprinting/republishing this material for advertising or promotional purposes, creating new collective works, for resale or redistribution to servers or lists, or reuse of any copyrighted component of this work in other works.}
}

\DeclareMathAlphabet{\mathcal}{OMS}{cmsy}{m}{n}

\title{\LARGE \bf
RaNDT SLAM: Radar SLAM Based on Intensity-Augmented Normal Distributions Transform
}

\author{Maximilian Hilger$^{1,2}$, Nils Mandischer$^{1,3}$, and Burkhard Corves$^{1}$
\thanks{$^{1}$All authors are with the Institute of Mechanism Theory, Machine Dynamics and Robotics, RWTH Aachen University, Aachen, Germany.}%
\thanks{$^{2}$Maximilian Hilger is also with the Chair of Perception for Intelligent Systems, Munich Institute of Robotics and Machine Intelligence, TU Munich, Munich, Germany. {\tt\small maximilian.hilger@tum.de}}%
\thanks{$^{3}$N. Mandischer is also with the Chair of Mechatronics, University of Augsburg, Augsburg, Germany. {\tt\small nils.mandischer@uni-a.de}}%
\thanks{This work was funded by the BMBF under grant ID 13N16482.}
}

\begin{document}

\maketitle
\thispagestyle{empty}
\pagestyle{empty}
\thispagestyle{specialfooter}

\begin{abstract}
Rescue robotics sets high requirements to perception algorithms due to the unstructured and potentially vision-denied environments. Pivoting Frequency-Modulated Continuous Wave radars are an emerging sensing modality for SLAM in this kind of environment. However, the complex noise characteristics of radar SLAM makes, particularly indoor, applications computationally demanding and slow. In this work, we introduce a novel radar SLAM framework, RaNDT~SLAM, that operates fast and generates accurate robot trajectories. The method is based on the Normal Distributions Transform augmented by radar intensity measures. Motion estimation is based on fusion of motion model, IMU data, and registration of the intensity-augmented Normal Distributions Transform. We evaluate RaNDT~SLAM in a new benchmark dataset and the Oxford Radar RobotCar dataset. The new dataset contains indoor and outdoor environments besides multiple sensing modalities (LiDAR, radar, and IMU).
\end{abstract}

\section{INTRODUCTION}
Radar-based Simultaneous Localization and Map-ping~(SLAM) is an emerging alternative to optical sensors in vision-restricted environments. Compared to LiDAR, the long wavelength in the millimeter range of Frequency-Modulated Continuous Wave~(FMCW) radars can easily penetrate fog, dust, and similar volumetric disturbances~\cite{Fritsche.2018}. Applications and existing approaches may be loosely grouped into the domains of autonomous driving~\cite{Checchin.2010,Schuster.2016,Hong.2022,Wang.25.10.2022,Adolfsson.2023} and rescue robotics~\cite{Fritsche.2018,.2013c,M.Mielle.2019,Torchalla.2021}. They offer different environmental characteristics. The confined indoor environments typically encountered by rescue robots lead to high noise in the radar signals. Therefore, significant computational effort has to be spent dealing with this noise, which leads to slow update rates. Outdoor radar SLAM approaches (e.g., autonomous driving) can rely on simpler techniques but are often untested in indoor environments. In this work, we focus on rescue robotics. Note, that rescue robots not only operate indoors, but can also be deployed in outdoor disaster scenarios, hence, we need to address two challenges: real-time performance and applicability in multiple environments. To this end, we propose a novel SLAM framework, RaNDT~SLAM, based on scan registration using the Normal Distributions Transform~(NDT), which takes not only the point entities but also the intensity of the radar signal into account. We use submaps storing intensity-augmented NDTs to accumulate filtered radar data. We estimate motion locally by fusing NDT registration, the yaw component of IMU measurements, and a motion model. Global consistency is ensured using a pose graph and a loop closure detector.

Our main contributions in RaNDT~SLAM are:
\begin{itemize}
	\item Novel modular estimation framework using radar scans, IMU measurements, and a motion model.
	\item Demonstration of real-time performance of our system in indoor and outdoor environments using a new benchmark dataset and the Oxford Radar RobotCar dataset\footnote[4]{We will use the shortened term "Oxford dataset" in the following.}.
	\item Implementation\footnote[5]{\url{https://github.com/IGMR-RWTH/RaNDT-SLAM}\label{ft:2}} and dataset~\cite{HM23} publicly available, providing a new baseline for radar SLAM in rescue robotics. To the authors' knowledge, this is the first dataset for SLAM using indurad's iSDR-C radar sensor.
\end{itemize}

In the following, we first briefly review existing radar SLAM approaches using pivoting radar sensors and introduce the NDT and its applications in radar scan registration (Section~\ref{sec:sota}). In Section~\ref{sec:methodology}, we introduce and explain RaNDT~SLAM. We evaluate the proposed framework profoundly in indoor and outdoor scenarios, and show its real-time capability (Section~\ref{sec:evaluation}). Finally, the results are summarized and potential additions are discussed (Section~\ref{sec:conclusion}). 

\section{RELATED WORK}
\label{sec:sota}
Pivoting FMCW radars are a commodity for SLAM in rescue robotics. We introduce SLAM algorithms using FMCW radars and the NDT SLAM, which forms this work's base.
\begin{figure*}[t!]
    \centering
	\includegraphics[width = .72\textwidth]{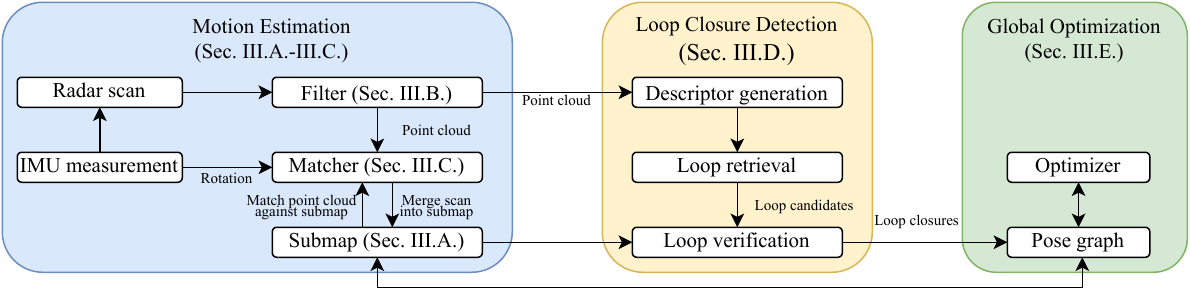}
    \hfill
    \includegraphics[width = .23\textwidth]{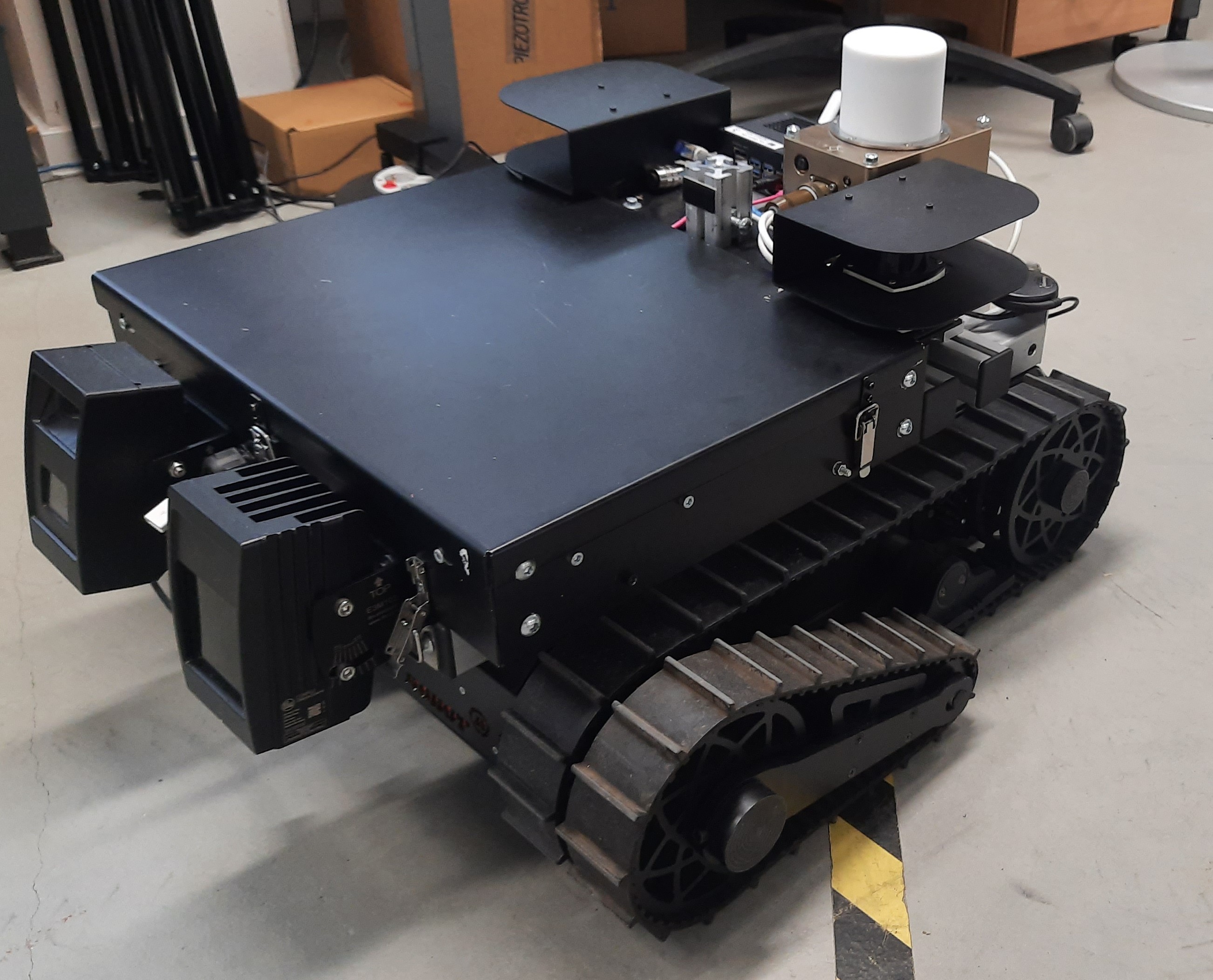}    
	\caption{Architecture of the RaNDT~SLAM framework and robot used for evaluation in Section~\ref{ssec:randt_slam_data}. The robot is equipped with an indurad iSDR-c pivoting radar, a phidgets IMU, and a pair of Sick TiM LiDARs.}
	\label{fig:framework}
\end{figure*}
 
\subsection{Related Work with Pivoting Radar Sensors}
We categorize the existing radar SLAM systems based on their target environment into indoor and outdoor:

\subsubsection{Indoor Environments}
The radar sensor used for the indoor SLAM of Marck et al.~\cite{.2013c} generates scans using only the strongest return per azimuth. They estimate motion based on scan-to-scan ICP and generate a map using a grid-mapping Particle Filter. In the \emph{SmokeBot} project, a mechani\-cally pivoting radar (MPR) was developed~\cite{Fritsche2016}. Within the project, Fritsche et al.~\cite{Fritsche.2018} combine radar and LiDAR scans for SLAM and investigate two methods: a feature-based approach using a Kalman filter, and a scan registration-based method. Further, Mielle, Magnusson, and Lilienthal~\cite{M.Mielle.2019} compare \emph{gmapping} and NDT-OM using only radar data produced by the MPR. Mandischer et al.~\cite{N.Mandischer.2020} pre-process radar scans using Otsu thresholding and beam-wise clustering. The motion is estimated by probabilistic iterative correspondences matching and an autoregression-moving-average filter for outlier rejection. Torchalla et al.~\cite{Torchalla.2021} propose a custom sensor setup using a pivoting single-chip radar. They use Truncated Signed Distance Functions for map representation and scan registration.  

\subsubsection{Outdoor Environments}
For outdoor deployment, Hong et al.~\cite{Hong.2022} detect blob features in Cartesian radar images and track them with a Kanade-Lucas-Tomasi tracker. They estimate motion by matching features between scans and solving them for the transform via Singular Value Decomposition. Loop closures are detected using the \emph{M2DP} descriptor. Wang et al.~\cite{Wang.25.10.2022} find linear features in filtered point clouds. They use scan-to-map matching based on a point-to-line metric. Features that appear infrequently are erased from the map. Loop closures are detected using the \emph{ScanContext} Descriptor~\cite{Kim.2022}. Adolfsson et al.~\cite{Adolfsson.2023} use the strongest returns in each azimuth to compute sparse oriented surface points. They estimate motion using a weighted point-to-point metric and detect loop closures using a combination of \emph{ScanContext}, an odometry similarity measure, and alignment quality metrics. 

\subsection{Normal Distributions Transform SLAM}
The NDT subdivides a scan using a regular grid. All points within a grid cell are approximated by a normal distribution with mean $\bm{\mu}$ and covariance $\bm{\Sigma}$ defined as
\begin{equation}
\bm{\mu} =
\begin{pmatrix}
\mu_x\\
\mu_y
\end{pmatrix},
\bm{\Sigma} =
\begin{pmatrix}
\Sigma_{xx} & \Sigma_{xy} \\
\Sigma_{yx} & \Sigma_{yy}  
\end{pmatrix}.
\label{eq:ndt_parameters}
\end{equation}
The indices $x$ and $y$ denote the geometrical dimensions. Two paradigms are used to find the affine transformation matrix
\begin{equation}
	\mathbf{T} =
	\begin{pmatrix}
		\mathbf{R} & \mathbf{t} \\
		\mathbf{0} & 1 
	\end{pmatrix},
\end{equation}
with rotation matrix $\mathbf{R}$ and translation vector $\mathbf{t}$ that aligns a moving scan (MS) to a fixed scan (FS): Point-to-Distribution~(P2D) and Distribution-to-Distribution matching~(D2D). In P2D~\cite{Biber.2003}, the FS is converted into a NDT. Each point of the MS is evaluated on the NDT. The sum of all evaluated points is used as score function. The resulting transform is obtained by optimizing this score function. In D2D~\cite{Stoyanov.2012b}, the $L_{2}$ distance between NDT~$\mathcal{M}$ of the MS and NDT~$\mathcal{F}$ of the FS is minimized. This yields the cost function
\begin{equation}
	C_{d2d} = -d_1 \sum_{i=1}^{n_\mathcal{M}} \sum_{j=1}^{n_\mathcal{F}} \exp \left(-\frac{d_2}{2} \bm{\mu}_{ij}^T(\bm{\Sigma}_i + \mathbf{R} \bm{\Sigma}_j \mathbf{R}^T) \bm{\mu}_{ij}\right),
	\label{eq:d2dndt}
\end{equation}
where $d_1$ and $d_2$ are positive regularization parameters, and $\bm{\mu}_{ij} = \mathbf{R}\bm{\mu}_i + \mathbf{t} - \bm{\mu}_j$ is a short notation for the difference of the means of two distributions. The number of elements in the moving and fixed NDT are given by~$n_\mathcal{M}$ and~$n_\mathcal{F}$, respectively. While developed for LiDAR, the NDT is frequently used for radar scan registration. Some approaches aim to use the radar's intensity information to improve registration accuracy. Heller, Petrov, and Yarovoy~\cite{Heller.2022} use the intensity to weight each point in both, distribution calculation and alignment optimization. Kung, Wang, and Lin~\cite{Kung.2021} extend their approach by shifting the returned power, further increasing the influence of high-intensity points. In addition, the covariance of each cell is set depending on the distance of the points to the sensor. Our approach is most similar to~\cite{Heller.2022}, but by using the intensity as an additional dimension in the normal distributions, we aim to increase the robustness, particularly in indoor environments.

\section{METHODOLOGY}
\label{sec:methodology}
In the following, we propose the framework \textbf{Ra}dar \textbf{N}ormal \textbf{D}istributions \textbf{T}ransform~(RaNDT)~SLAM. The information is organized in three hierarchical layers: states, keyframes, and submaps. Every scan at time~$t$ is associated with a state $\mathbf{x}_{t} = \{\mathbf{T}_t, \mathbf{v}_t, b_t\}$, with the 2D pose~$\mathbf{T}_{t} \in SE(2)$ given in the submap frame, the velocity~$\mathbf{v}_{t} \in \mathbb{R}^3$ given in the robot frame, and an IMU bias~$b_{t}$. Each scan is filtered and subsequently used for local motion estimation. For further processing, we select keyframes. These are used to detect loop closures whose poses~$\mathbf{P}_k$ form the nodes of the pose graph. Lastly, small submaps~$\mathcal{F}$ are used to accumulate the measurements of the keyframes. This overcomes the possible sparseness of individual scans. An overview of the frameworks' architecture is given in Figure~\ref{fig:framework}.

\subsection{Map Representation}
\label{subsec:map_representation}
The map generated by our framework is a collection of small submaps. These are represented as NDTs. In contrast to the aforementioned approaches (see Section~\ref{sec:sota}), we augment the distribution's parameters introduced in~Equation~\ref{eq:ndt_parameters} with an intensity dimension $p$. The augmented parameters are 
\begin{equation}
	\tilde{\bm{\mu}} =
	\begin{pmatrix}
		\mu_x\\
		\mu_y\\
		\mu_p
	\end{pmatrix},
	\tilde{\bm{\Sigma}} =
	\begin{pmatrix}
		\Sigma_{xx} & \Sigma_{xy} & \Sigma_{xp} \\
		\Sigma_{yx} & \Sigma_{yy} & \Sigma_{yp} \\
		\Sigma_{px} & \Sigma_{py} & \Sigma_{pp}
	\end{pmatrix}.
\end{equation}
To obtain the distributions, we use the sample mean and covariance of all intensity-augmented points $\tilde{\mathbf{p}} = [p_x~p_y~p_p]^T$ falling into a cell as proposed by Magnusson et al.~\cite{Magnusson.2007}. With the additional intensity value in the augmented mean vector~$\tilde{\bm{\mu}}$, incorrect correspondences with largely differing intensities can be suppressed: large values inside the exponent of Equation~\ref{eq:d2dndt} are treated as outliers and have no influence in the optimization.
The augmented covariance matrix~$\tilde{\bm{\Sigma}}$ captures not only the geometric distribution of the points, but expresses also the covariance between the intensity and the geometric coordinates. Hence, the covariance implicitly models the target material's reflection characteristics. Using both, augmented mean and covariance, we aim to improve the registration accuracy in ambiguous indoor applications, where materials reflect with different intensities. The submaps are built using the scans of all keyframes. New sensor data is integrated using the recursive update formulation by Saarinen et al.~\cite{J.Saarinen.2013}. As the submaps only contain a small part of the environment, we do not explicitly track occupancy values within each cell.

\subsection{Radar Filtering}
\label{subsec:filtering}
The indoor application of our framework makes aggressive radar filtering necessary. We propose using a two-step procedure. First, we discard low-intensity and far/close points. Afterwards, we extract the cluster around the most intense point of each radar beam. For a point to belong to the cluster, two requirements have to be fulfilled:
\begin{itemize}
	\item The point has a smaller distance than a threshold to another point of the cluster.
	\item The point's intensity is lower than the intensity of the neighboring point closer to the point of highest intensity.
\end{itemize}
The choice of the intensity threshold should reflect the target environment and the sensor specifics. In indoor applications, higher values can suppress the stronger multi-path reflections. Outdoors, the threshold can be set lower. This efficiently reduces the scan size of the radar from initially approx. $10000$ to $1500$~points. While this aggressive filtering approach is susceptible to missing important secondary landmarks, our experiments show that competitive motion estimation accuracy performance can be achieved. The radar point cloud before and after filtering is depicted in Figure~\ref{fig:filter}.

\subsection{Motion Estimation}
\label{subsec:local_motion_estimation}
\begin{figure}[t!]
	\centering
	\includegraphics[width = 0.48\textwidth]{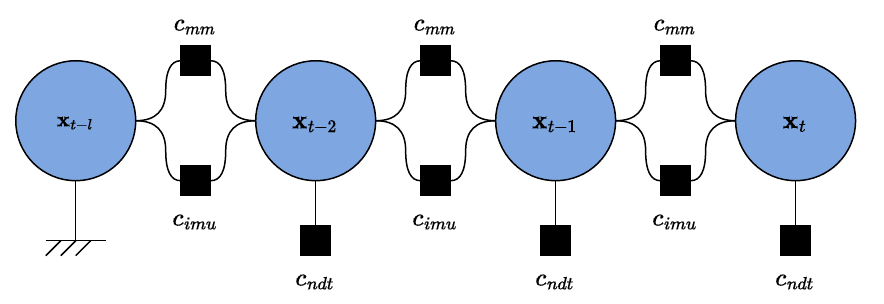}
	\caption[Factor graph optimizing over $l=3$ states]{Factor graph optimizing over $l=3$ states.}
	\label{fig:factor_graph}
\end{figure}

Local Motion estimation is necessary to build the submap and obtain an accurate estimate of the last state. Our state estimator uses three components to pose a least squares optimization problem (according indices annotated):
\begin{itemize}
	\item Motion model ($mm$)
	\item Relative rotations obtained by an IMU ($imu$)
	\item NDT registration between scan and submap ($ndt$)
\end{itemize}
We explicitly not use wheel odometry as rescue robots usually employ a tracked drive train, leading to poor odometry. All components are jointly optimized using the cost function
\begin{equation}
	C(\mathbf{x}_{t_{l}:t}) = \sum_{t_{l}}^{t} C_{ndt}(\mathbf{x}_{i}) + C_{mm} (\mathbf{x}_{i-1},\mathbf{x}_{i}) + C_{imu}(\mathbf{x}_{i-1}, \mathbf{x}_{i}),
\end{equation}
where $t_{l}=t-l+1$. The cost function is optimized over~$l$ time steps. This optimization over a window of states forces consistency over multiple poses. In ambiguous situations, this can prevent incorrect convergence. New scans are inserted into the map after leaving the optimization window. Selecting $l$ is a trade-off between registration accuracy and computational efficiency. We experimentally determine that $l=3$ works well across multiple domains. As $l$ also determines the delay before scan insertion, an improper value can lead to insufficient scan overlap, impairing the registration. Figure~\ref{fig:factor_graph} depicts the factor graph corresponding to the cost function. Following, the individual components are introduced:
\begin{figure}[t!]
	\centering
	\begin{subfigure}[b]{0.27\textwidth}
		\includegraphics[width = 1\textwidth]{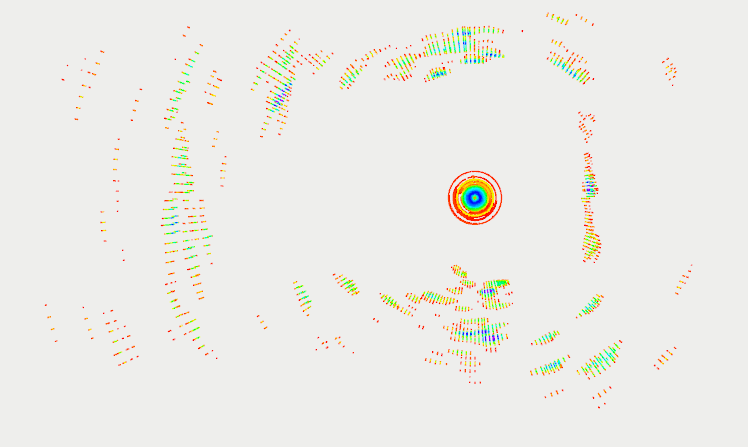}
		\caption{Raw radar scan (indurad iSDR-C).}
		\label{fig:unfiltered}
	\end{subfigure}
	\hspace*{\fill}
	\begin{subfigure}[b]{0.193\textwidth}
		\includegraphics[width = 1\textwidth]{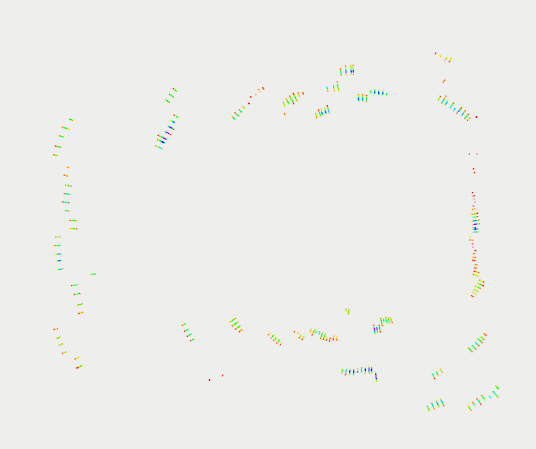}
		\caption{Filtered radar scan.}
		\label{fig:filtered}
	\end{subfigure}
	\caption[Effect of the radar filter]{Effect of the filter on the radar scan. Intensities are denoted red/low to purple/high; radome is circular entity in~\subref{fig:unfiltered}.}
	\label{fig:filter}
\end{figure}

\subsubsection{Motion Model}
The motion model generates the pose prediction
\begin{equation}
  \hat{\mathbf{T}}_{t} = \mathbf{T}_{t-1} \operatorname*{Exp}(\mathbf{v}_{t-1} \Delta t).
\end{equation}
We use the notation in~\cite{Sola.04.12.2018}, where \mbox{$\operatorname*{Exp}(\cdot),\, \mathbb{R}^3 \to SE(2)$} denotes the exponential map, lifting an element of the tangent space to the group element. Accordingly, the logarithmic map \mbox{$\operatorname*{Log}(\cdot),\,  SE(2) \to \mathbb{R}^3$} gives the tangent vector that corresponds to the group element. The motion cost is obtained as the Mahalanobis distance between prediction and estimation 
\begin{equation}
	C_{mm} = \left\lVert
	\begin{pmatrix}
		\operatorname*{Log}(\hat{\mathbf{T}}_t^{-1} \mathbf{T}_t) \\
		\mathbf{v}_t - \mathbf{v}_{t-1}
	\end{pmatrix}
	\right\rVert_{\bm{\Omega}_{mm}}^2,
\end{equation}
where $\bm{\Omega}_{mm}$ is the information matrix of the motion model. The values of $\bm{\Omega}_{mm}$ are selected empirically.

\subsubsection{IMU Data}
The measurements of the IMU are fused as a relative rotation~$\Delta\mathbf{R}_{t}$, computed by a complementary filter~\cite{Valenti.2015}. The bias accounts for the drift of the angular velocity measured by the IMU's gyroscope. This gives the rotational cost function
\begin{equation}
	C_{imu} = 
	\left\lVert
	\begin{pmatrix}
		\operatorname*{Log}\left((\mathbf{R}_{t-1} \Delta \mathbf{R}_t \operatorname*{Exp}(b_{t-1} \Delta t))^{-1} \mathbf{R}_t\right) \\
		b_{t} - b_{t-1} 
	\end{pmatrix}
	\right\rVert_{\bm{\Omega}_{imu}}^2,
\end{equation}
where $\bm{\Omega}_{imu}$ is the information matrix of the IMU model.

\subsubsection{NDT Registration}
Two adjustments have to be made to integrate the NDT into the estimator. First, the parameters to optimize for ($\mathbf{R}$ and $\mathbf{t}$) have to be adjusted to be compliant with the augmented NDT cells. The augmented rotation matrix and the modified translation vector are given by
\begin{equation}
	\tilde{\mathbf{R}} =
	\begin{pmatrix}
		\mathbf{R} & \mathbf{0}\\
		\mathbf{0} & 1 
	\end{pmatrix},
	\tilde{\mathbf{t}} =
	\begin{pmatrix}
		\mathbf{t} \\
		0
	\end{pmatrix}.
\end{equation}
This results in the augmented D2D-NDT cost function between the scan's NDT~$\mathcal{M}$ and submap~$\mathcal{F}$, given by 
\begin{equation}
	C_{d2d} = -d_1 \sum_{i=1}^{n_\mathcal{M}} \sum_{j=1}^{n_\mathcal{F}} \exp \left( -\frac{d_2}{2} \tilde{\bm{\mu}}_{ij}^T \left(\tilde{\bm{\Sigma}}_i + \tilde{\mathbf{R}} \tilde{\bm{\Sigma}}_j \tilde{\mathbf{R}}^T \right) \tilde{\bm{\mu}}_{ij} \right) \, .
	\label{eq:d2dndt aug}
\end{equation}
Commonly, this function is optimized using Newton's method~\cite{Stoyanov.2012b}. To express this in a least squares formulation, the second adjustment has to be made. Note, that the NDT cost function has the same shape and gradient as the Welsch loss~\cite{Barron.2019}
\begin{equation}
	\rho(r^2, c) =  1 - \exp\left(-\frac{r^2}{2 c^2}\right),
\end{equation}
with an offset of 1. Therefore, the NDT cost can equally be described by the sum of multiple quadratic error terms
\begin{equation}
r^{2} = \tilde{\bm{\mu}}_{ij}^T \left(\tilde{\bm{\Sigma}}_i + \tilde{\mathbf{R}} \tilde{\bm{\Sigma}}_j \tilde{\mathbf{R}}^T \right) \tilde{\bm{\mu}}_{ij}
\end{equation}
and a robust loss function $\rho(r^2, c)$, where $c$ sets the scale of the loss function and $r$ denotes the residual. Also other robust loss functions can be used. Each loss can be interpreted as resulting from a specific outlier distribution. Adaptive loss functions can interpolate between multiple variants and, therefore, allow for tailored tuning on different environment properties~\cite{Barron.2019}.

\subsubsection{Adaptive Loss Formulation}
To avoid falling into local minima, a frequently used approach is to register scans on multiple scales. In NDT matching, this is mostly accomplished by matching different map resolutions~\cite{J.Saarinen.2013}. A different approach in scan matching is to adapt the scale~$c$ of the optimization problem. For a macroscopic scale, the objective function is easy to optimize, but may be affected by outliers. With increasingly smaller scales, the influence of outliers vanishes at the cost of more local minima. By solving a series of optimization problems with decreasing~$c$, the risk of incorrect convergence can be mitigated~\cite{Yang.2020}. To leverage both, the adaptive loss formulation and the multi-scale optimization, we use the adaptive loss 
\begin{equation}
	\rho(r^2, \alpha, c, \mu) = \frac{|\alpha -2|}{\alpha} \left(\left( \frac{(r^2/(\mu c^2))}{|\alpha-2|}+1 \right)^{\frac{\alpha}{2}} -1 \right).
\end{equation}
This formulation is derived from~\cite{Barron.2019}, where the shape parameter~$\alpha$ is introduced to interpolate different loss functions. In addition, the parameter~$\mu$ is used to adjust the scale. In each iteration~$n$, the parameter is reduced using \mbox{$\mu_{n+1} \gets \mu_n / k_\mu$}. In the final iteration, the control parameter reaches $\mu=1$.
 
\subsubsection{Resulting NDT Cost Function}
The double sum in Equation~\ref{eq:d2dndt aug} is computationally expensive, hence, only a subset of closest correspondences is considered. For each normal distribution in the scan's NDT, we select the four closest distributions in the submap. This yields the set of correspondences~$\mathcal{K}$. The resulting cost function is given by
\begin{equation}
	C_{ndt} = \frac{ w_{ndt}}{n_\mathcal{K}} \sum_{(i,j) \in \mathcal{K}} \rho \left(\tilde{\bm{\mu}}_{ij}^T \left(\tilde{\bm{\Sigma}}_i + \tilde{\mathbf{R}} \tilde{\bm{\Sigma}}_j \tilde{\mathbf{R}}^T \right) \tilde{\bm{\mu}}_{ij}, \alpha, c, \mu \right),
\end{equation}
where $w_{ndt}$ is empirically tuned to set the influence of the NDT in the optimization. The final solution is obtained using \emph{Ceres}~\cite{.2012} and the Levenberg-Marquardt algorithm.

\subsection{Loop Closure Detection}
\label{subsec:loop_closure}
For loop closure detection, we make use of the method proposed in~\cite{Adolfsson.2023}. Here, the filtered scan of each keyframe is used in its polar representation to generate the \emph{ScanContext} Descriptor~$\mathbf{I}$. The scan is binned in angular and radial direction. Each element of the descriptor is calculated as the sum of the intensities of all points divided by $20$ within the bin. When creating a new descriptor, a loop closure search is triggered. Given a query scan~$q$, the aim is to find the most likely loop candidate $lc$. This likelihood is expressed using a combination of the descriptor distance~$d_{sc}(\mathbf{I}^q, \mathbf{I}^{lc})$ and an odometry similarity~$d_{od}^{q,lc}$, given by
\begin{equation}
	lc = \operatorname*{argmin}_{lc} \left(d_{sc}(\mathbf{I}^q, \mathbf{I}^{lc}) + d_{od}^{q,lc}\right).
\end{equation}
The highest-scoring loop candidate is further refined by matching the query scan with the submap containing the loop candidate. After alignment, we use the Cauchy-Schwarz divergence between the NDTs of both, query scan and the loop candidate's submap, as gate to accept or reject the loop closure as described in~\cite{Tabib.2021}. This divergence between two probability density functions~$p(x)$ and~$q(x)$ is given by 
\begin{equation}
	D_{CS}(q,p) = -\log \left( \frac{\int q(x)p(x) dx}{\sqrt{\int q(x)^2 dx \int p(x)^2 dx}}\right),
\end{equation} 
which can be expressed in closed form for Gaussian Mixture Models such as the NDT~\cite{Kampa.2011}. If the divergence exceeds a threshold, the loop closure is discarded. Otherwise, a constraint between the query node and the root node of the loop candidate's submap is added to the pose graph.

\subsection{Global Optimization}
\label{subsec:pose_graph}
Global consistency is ensured by regularly optimizing the pose graph. The graph consists of keyframe poses~$\mathbf{P}$, odometry constraints~$\mathcal{C}_{od}$, and loop constraints~$\bm{\Omega}_{lo}$. To obtain the optimal trajectory, the cost function
\begin{equation}
	C_{pg}(\mathbf{P}) = \sum_{(a,b) \in \mathcal{C}_{od}} \left(\mathbf{e}_{ab}^T \bm{\Omega}_{od} \mathbf{e}_{ab}\right) + \sum_{(a,b) \in \mathcal{C}_{lo}} \left(\mathbf{e}_{ab}^T \bm{\Omega}_{lo} \mathbf{e}_{ab}\right)
\end{equation}
is minimized using the Levenberg-Marquardt algorithm. Here, the information matrices~$\bm{\Omega}_{od}$ and $\bm{\Omega}_{lo}$ are set to constant values. The pose errors~$\mathbf{e}_{ab}$ quantify the difference between the relative transform between the pose graph nodes and the relative transform stored in the constraint.

\section{EVALUATION}
\label{sec:evaluation}
We evaluate the performance of our approach in two new benchmark scenarios~\cite{HM23} to demonstrate the feasibility in varying environments. We discuss the most important parameters and their configuration. The resulting trajectories are benchmarked\footnote[6]{All experiments are performed on a PC running Ubuntu 20.04 in Docker with a Ryzen 2600X CPU at 3.6~GHZ and 16~GB RAM, without GPU.} against LiDAR SLAM. We compare our framework to state-of-the-art radar odometry and SLAM approaches based on the Oxford dataset~\cite{Barnes.2020b}. To evaluate RaNDT~SLAM, the estimated trajectory $\mathbf{P}_1,...,\mathbf{P}_n \in SE(2)$ is compared to a ground truth trajectory~$\mathbf{Q}_1,...,\mathbf{Q}_n \in SE(2)$. We evaluate two performance metrics: Relative Pose Error~(RPE) and Absolute Trajectory Error~(ATE) \cite{Sturm.2012}. In line with~\cite{Torchalla.2021}, we use the mean variant of the RPE. For the ATE, we use the Root Mean Squared Error~(RMSE) of the translation errors between the estimated and the ground truth poses:
\begin{subequations}
	\begin{align}
		\textit{T-RPE} &= \frac{1}{n-1} \sum\nolimits_{i=1}^{n-1} ||\textit{trans}(\mathbf{E}_i)||, \\
		\textit{R-RPE} &= \frac{1}{n-1} \sum\nolimits_{i=1}^{n-1} ||\textit{rot}(\mathbf{E}_i)||, \\
        \textit{ATE} &= \sqrt{\frac{1}{n} \sum\nolimits_{i=1}^{n} ||\textit{trans}(\mathbf{F}_i)||^2},
	\end{align}
\end{subequations}
with the translational ($\textit{trans}(\cdot)$) and a rotational ($\textit{rot}(\cdot)$) component, the mean of all individual errors~$\mathbf{E}_i$, and \mbox{$\mathbf{F}_i = \mathbf{Q}_i^{-1} \mathbf{P}_{i}$}. The poses of the ground truth trajectory are obtained at a different rate than the estimated poses. For every ground truth pose, we interpolate linearly between the two closest estimated poses. For the evaluation on the Oxford dataset (Section~\ref{ssec:oxford_eval}), we, additionally, employ the KITTI odometry error~\cite{Geiger.2012}. This metric measures the drift by computing the translation error and rotation error over trajectory segments of length \mbox{100, 200, ... , 800}~m. The drift is obtained by averaging over all evaluated segments. This metric is not applicable to the RaNDT~SLAM benchmark, as the trajectories are not sufficiently long.

\subsection{RaNDT SLAM Dataset}
\label{ssec:randt_slam_data}
\begin{figure}[t!]
	\begin{center}
    	\begin{subfigure}[b]{0.4\textwidth}
    		\includegraphics[width=\textwidth,]{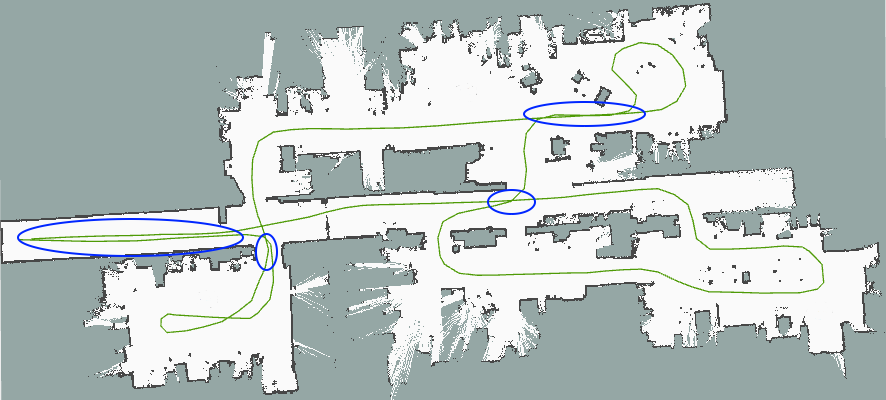}
    		\caption{Indoor environment.}
    		\label{fig:scenario1}
    	\end{subfigure}
    	\begin{subfigure}[b]{0.32\textwidth}
    		\includegraphics[width=\textwidth]{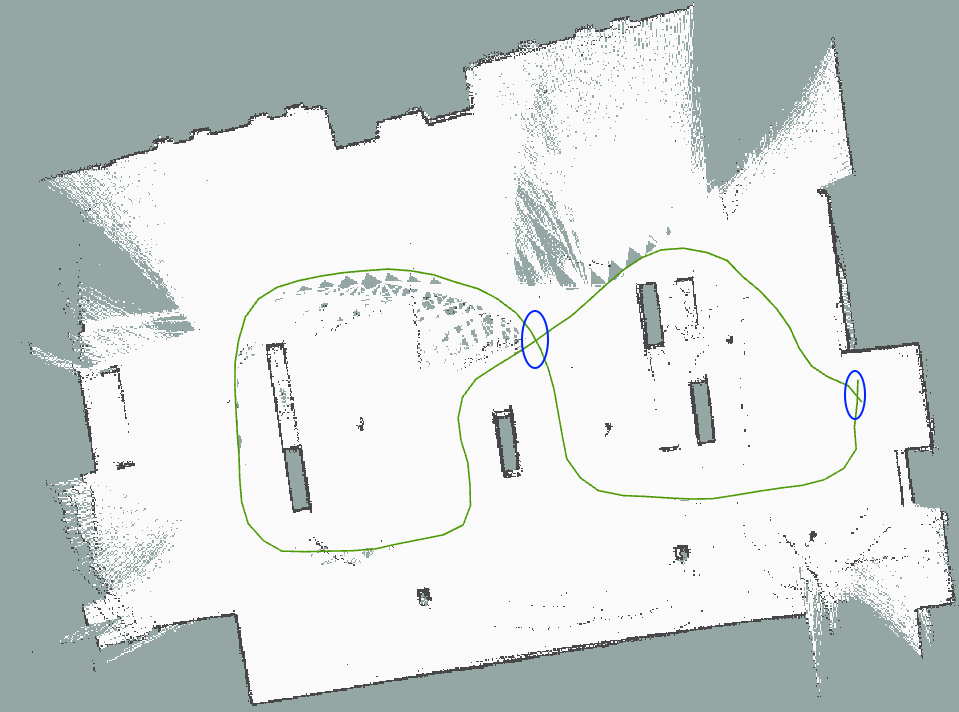}
    		\caption{Outdoor environment.}
    		\label{fig:scenario2}
    	\end{subfigure}
	\end{center}
	\caption{RaNDT SLAM scenarios (green: LiDAR trajectory, blue: potential loop closures); maps generated using LiDAR.}
	\label{fig:gt_map}
\end{figure}

The first scenario in the RaNDT~SLAM dataset is recorded in a lab space (Figure~\ref{fig:scenario1}). The test environment contains a long corridor, two labs, and a cluttered workshop. Multiple loops occur in the trajectory. The total trajectory is about $137$~m long and is traversed in $308$~s. The low velocity accounts for the expected slow operating speed in rescue robotics applications. To assess the outdoor performance, a second scenario is recorded in a sparse courtyard (Figure~\ref{fig:scenario2}). The robot drives in an eight-like shape, containing two possible loop closures. Compared to the indoor scenario, the outdoor setting offers less features and has varying inclination angles. Latter violates the 2D assumption and further tests the robustness. All scenarios are recorded using a tracked robot equipped with an iSDR-C radar sensor by indurad and a Phidgets Spatial IMU. The radar sensor offers a range resolution of $16.1$~mm and an angular resolution of $1.64$\textdegree. The radar spins at $5$~Hz and has a maximum range of $16$~m. The benchmark dataset is publicly available~\cite{HM23}.
\begin{figure*}[t!]
	\begin{center}
    	\begin{subfigure}[b]{0.582\textwidth}
    		\centering
    		\includegraphics[width=\textwidth]{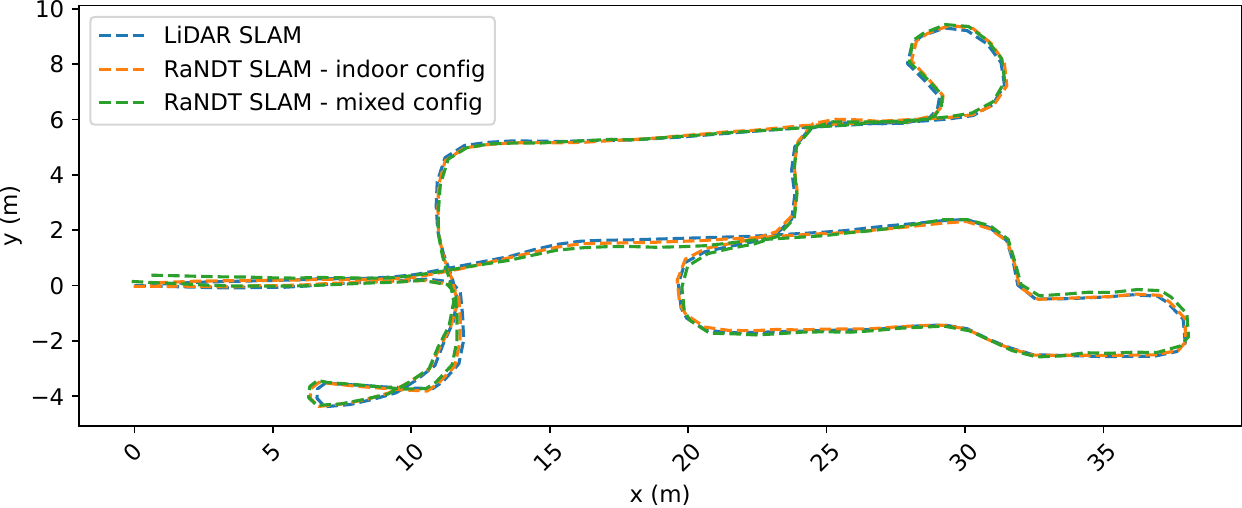}
    		\caption{Indoor environment.}
    		\label{fig:indoor_combined}
    	\end{subfigure}
        \begin{subfigure}[b]{0.39\textwidth}
    		\centering
    		\includegraphics[width=\textwidth]{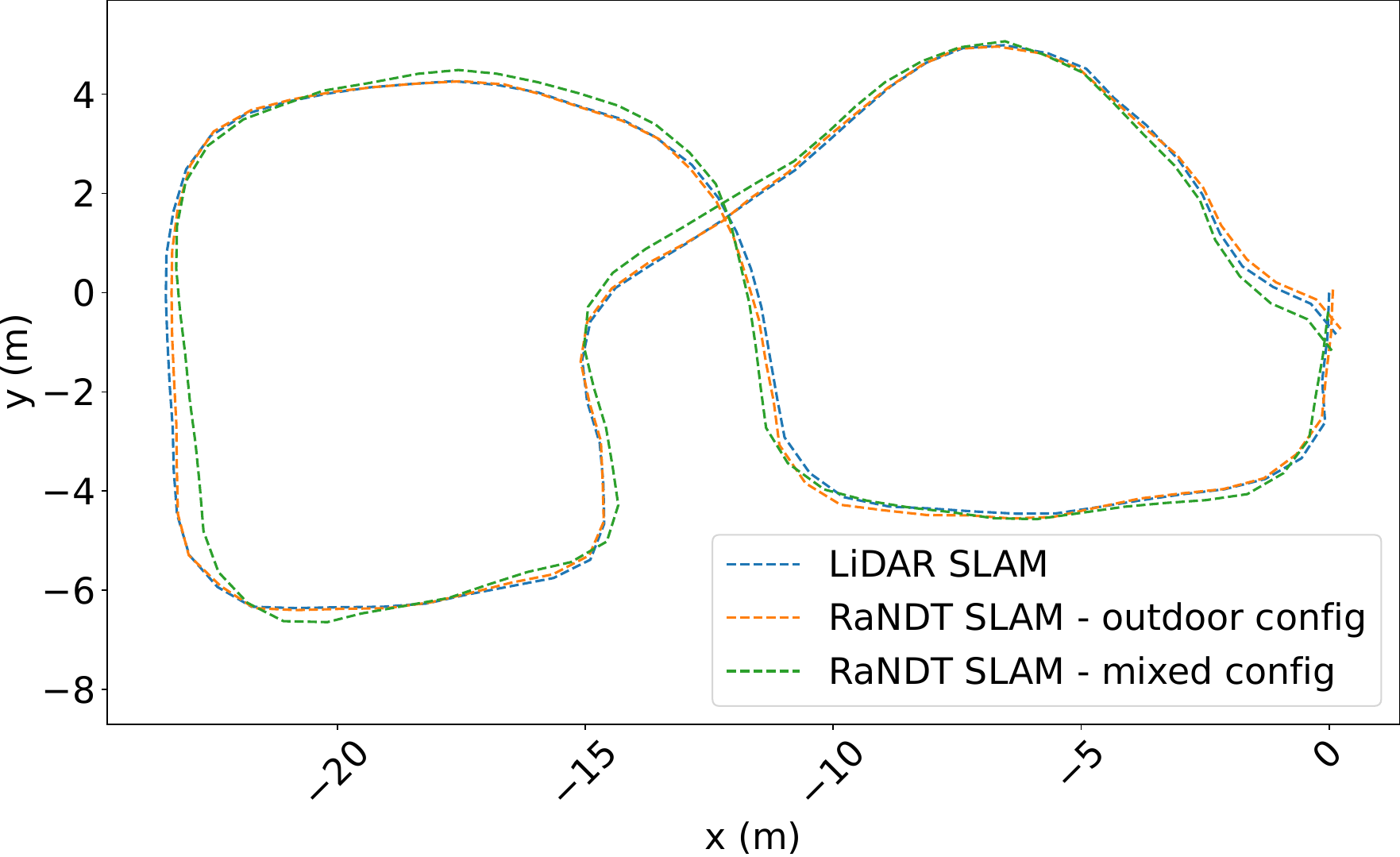}
    		\caption{Outdoor environment.}
    		\label{fig:outdoor_combined}
    	\end{subfigure}
	\end{center}
	\caption{Estimated trajectories in different environments, using specific and mixed parameter configurations.}
	\label{fig:randt_trajectories}
\end{figure*}

\subsection{Parameters}
\label{ssec:eval_parameters}
We prepare three parameter configurations using an extensive parameter study: A specialized configuration for each, indoor and outdoor, application, and a mixed configuration suitable for both. Most prominently, the resolution of the map has to be set properly for the outdoor application. Indoors, a resolution of $0.5$~m yields great results, but fails outdoors. Hence, the outdoor resolution is set to $1.2$~m. The mixed configuration uses a resolution of $1$~m to trade off accuracy in both, indoor and outdoor environment. In addition, we observe that the outdoor configuration requires a larger shape~$\alpha$ and a larger scale~$c$ compared to the indoor one. This is in line with the expected higher outlier ratio indoors produced by stronger multi-path reflections. For indoor usage, we use $\alpha=-2.0$ and $c=1.5$. Outdoors, we choose $\alpha=-1.0$ and $c=2.0$. The mixed configuration is set to $\alpha=-1.5$ and $c=2.0$. The complete parameter configurations may be found in our code publication of RaNDT~SLAM\footref{ft:2}. While the parameter settings for the Oxford dataset are based on the outdoor configuration, several adjustments need to be made: Firstly, the intensity threshold is set to $p=70$, as the Navtech sensor returns higher intensity values. In addition, we set the resolution to $3.5$~m to account for the larger scale of the environment. Finally, we reduce the number of scans that are inserted into a submap. This accounts for the higher velocity of the robot in autonomous driving scenarios.

\subsection{Evaluation with RaNDT Radar Benchmark}
Within our new dataset, we use \emph{SlamToolbox}~\cite{Macenski.2021} with LiDAR data to generate a ground truth trajectory. The comparison of radar with LiDAR trajectories is commonly applied in absence of more accurate ground truth measures, see, e.g.,~\cite{Torchalla.2021}. However, it should be noted that the state of the art in radar SLAM approaches the accuracy of LiDAR SLAM~\cite{Adolfsson.2023}. We perform ten runs for each scenario and in each parameter configuration. This accounts for slight variations in the performance measures due to the multi-threaded implementation. The indoor environment is tested with the indoor and mixed configuration, and the outdoor environment with the outdoor and mixed configuration. Using the indoor configuration on the outdoor environment and vice versa proofed to be infeasible (compare Section~\ref{ssec:eval_parameters}). In addition, we evaluate the specialized configurations without loop closure to demonstrate the improved performance enabled by the loop closure module. Table~\ref{tab:performance} lists ATE and RPE measures of the tests using the LiDAR trajectory as ground truth.

\subsubsection{Indoor}
In the indoor scenario, both, the indoor and the mixed configuration, show accurate performance (Figure~\ref{fig:indoor_combined}). Regarding ATE and R-RPE, the mixed configuration performs only slightly worse. The T-RPE increases by $18.8$\%, indicating that the local accuracy deteriorates. Here it can be observed, that the trajectory obtained with the mixed configuration deviates slightly from ground truth on long straights. However, we conclude that both configurations are suitable for indoor usage. Comparing the performance gain enabled by the loop closure, we observe an ATE reduction of 57\%, underlining the importance of the loop closure. The RPEs increase slightly.

\subsubsection{Outdoor}
The outdoor trajectories generated by outdoor and mixed configuration are depicted in Figure~\ref{fig:outdoor_combined}. The trajectory estimated by the outdoor configuration closely aligns to the ground truth trajectory. This is also reflected in the low ATE measure. When using the mixed configuration, the ATE increases significantly. We observe that the overall trajectory quality remains high, but the long straight (at $x\approx -23$~m) deviates. Hence, the mixed configuration deteriorates more in outdoor environments. However, as the pure outdoor configuration is infeasible in indoor environment and vice versa, the mixed configuration is a good choice in case of the robot switching between outdoor and indoor environments, e.g., in case of large scale disaster sites with multiple buildings. In the outdoor environment, we observe an ATE reduction by 32\%. The gain is less than indoors, which can be explained by the smaller amount of loop closures. In conclusion, RaNDT~SLAM indicates high accuracy in both, indoor and outdoor, environments. The loop closure module improves the global consistency of the estimated trajectory significantly. When targeting a specific environment or use-case scenario, the performance can be further enhanced by deploying the according parameter configuration.
\begin{table}[t!]
	\centering
	\caption{Performance metrics generated by RaNDT~SLAM with (w.) and without (w/o) loop closure (LC).}
	\begin{tabular}{l l c|c c|c c }
		\hline
		\multicolumn{3}{l|}{Environment} & \multicolumn{2}{c|}{Indoor} & \multicolumn{2}{c}{Outdoor}\\
		\multicolumn{3}{l|}{Parameter Config.} & Indoor & Mixed & Outdoor & Mixed \\
		\hline
		\multirow{3}{*}{\rotatebox[origin=c]{90}{w. LC}} & $\textit{ATE}$ & [m]           & $0.143995$ & $0.169776$ & $0.242701$ & $0.422452$\\
		~ & $\textit{T-RPE}$ & [m]           & $0.033777$ & $0.040120$ & $0.061772$ & $0.063413$\\
		~ & $\textit{R-RPE}$ & [\textdegree] & $1.346297$ & $1.390718$ & $1.776593$ & $1.925730$\\
		\hline
		\multirow{3}{*}{\rotatebox[origin=c]{90}{w/o LC}} & $\textit{ATE}$ & [m]           & $0.337302$ & $-$ & $0.354598$ & $-$\\
		~ & $\textit{T-RPE}$ & [m]           & $0.032732$ & $-$ & $0.059621$ & $-$\\
		~ & $\textit{R-RPE}$ & [\textdegree] & $1.314041$ & $-$ & $1.706450$ & $-$\\
		\hline
	\end{tabular}	
	\label{tab:performance}
\end{table}
\begin{table}[t!]
	\centering
	\caption{Cycle times of individual subsystems in RaNDT~SLAM (SCD: \emph{ScanContext} Descriptor).}
	\begin{tabular}{r|c|c}
		\hline 
		Subtask & Duration indoor [ms] & Duration outdoor [ms] \\
		\hline
		Pre-Processing     & $2.743 \pm 0.923$   & $1.448 \pm 0.043$\\
		Motion Estimation  & $35.671 \pm 15.964$ & $18.404 \pm 8.603$\\
		SCD Matching       & $1.118 \pm 0.385$   & $2.195 \pm 0.844$\\
		Loop Refinement    & $14.266 \pm 5.833$  & $7.214 \pm 11.484$\\
		Graph Optimization & $54.833 \pm 34.623$ & $12.660$\\
		\hline
	\end{tabular}	
	\label{tab:computational_costs}
\end{table} 

\begin{table*}[t!]
    \scriptsize
    \centering
    \caption{Evaluation on Oxford dataset compared to state of the art.}
    \begin{tabular}{m{.01\textwidth}m{.168\textwidth}|C{.06\textwidth}C{.06\textwidth}C{.06\textwidth}C{.06\textwidth}C{.06\textwidth}C{.06\textwidth}C{.06\textwidth}C{.06\textwidth}|C{.06\textwidth}}
        \hline
        ~ & \textbf{Method} & \verb|10-12-32| & \verb|16-13-09| & \verb|17-13-26| & \verb|18-14-14| & \verb|18-15-20| & \verb|10-11-46| & \verb|16-11-53| & \verb|18-14-46| & \textbf{Mean} \\
        \hline
        \hline
        ~ & ~ &  \multicolumn{9}{|c}{\textbf{Drift (in \% / (\textdegree/$100$~m)) on the Oxford dataset comparing SLAM and odometry methods.}} \\
        \hline
        \multirow{10}{*}{\rotatebox[origin=c]{90}{Odometry}} & Cen2019 \cite{Cen.2019}& - & - & - & - & - & - & - & - & 3.72/0.95 \\
        ~ & MC-RANSAC \cite{Burnett.2020} & - & - & - & - & - & - & - & - & 3.31/1.09 \\
        ~ & CC-means \cite{Aldera.2022}  & - & - & - & - & - & - & - & - & 2.53/0.82 \\
        ~ & Under the radar \cite{Barnes.2020}& - & - & - & - & - & - & - & - & 2.06/0.67 \\
        ~ & R\textsuperscript{3}O \cite{Lubanco.2023} & 1.89/0.56 & 1.95/0.60 & 2.09/0.58 & 2.06/0.61 & 2.18/0.65 &2.02/0.57 & 2.17/0.60 & 1.96/0.57 & 2.04/0.59 \\
        ~ & MbyM \cite{Barnes.2019} & - & - & - & - & - & - & - & - & 1.16/\textbf{0.30} \\
        ~ & HERO \cite{Burnett.2021}& 1.77/0.62 & 1.75/0.95 & 2.04/0.73 & 1.83/0.61 & 2.20/0.77 & 2.14/0.71 & 2.01/0.61 & 1.97/0.65 & 1.96/0.66 \\
        ~ & Kung \cite{Kung.2021} & - & - & - & - & - & - & - & - & 1.96/0.60 \\
        ~ & SDRO \cite{Zhang.2023}& 1.43/\textbf{0.32} & 1.58/\textbf{0.24} & 1.52/\textbf{0.27} & 1.58/\textbf{0.35} & 1.66/\textbf{0.24} & 1.63/\textbf{0.32} & 1.78/\textbf{0.34} & 1.68/\underline{0.38} & 1.65/\textbf{0.30}\\
        ~ & CFEAR-3 \cite{Adolfsson.2023b} & 1.20/0.36 & 1.24/0.40 & 1.23/0.39 & 1.35/0.42 & 1.24/0.41 & 1.22/0.39 & 1.39/0.40 & 1.39/0.44 & 1.28/0.40\\
        ~ & CFEAR-3-s50 \cite{Adolfsson.2023b} & \textbf{1.05}/\underline{0.34} & \textbf{1.08}/\underline{0.34} & \underline{1.07}/0.36 & \textbf{1.11}/\underline{0.37} & \textbf{1.03}/0.37 & \textbf{1.05}/0.36 & \textbf{1.18}/\underline{0.36} & \underline{1.11}/\textbf{0.36} & \textbf{1.09}/\underline{0.36} \\
        \hline
        \multirow{3}{*}{\rotatebox[origin=c]{90}{SLAM}} & RadarSLAM-Full \cite{Hong.2022} & 1.98/0.60 & 1.48/0.50 & 1.71/0.50 & 2.22/0.70 & 1.77/0.60 & 1.96/0.70 & 1.81/0.60 & 1.68/0.50 & 1.83/0.60\\
        ~ & MAROAM \cite{Wang.25.10.2022} & 1.63/0.46 & 1.83/0.56 & 1.49/0.47 & 1.54/0.47 & 1.61/0.50 & 1.55/0.53 & 1.78 /0.54 & 1.55/0.50 & 1.62/0.50 \\
        ~ & TBV Radar SLAM \cite{Adolfsson.2023} & \underline{1.17}/0.35 & \underline{1.15}/0.35 & \textbf{1.06}/\underline{0.35} & \underline{1.12}/\underline{0.37} & \underline{1.09}/\underline{0.36} & \underline{1.18}/\underline{0.35} & \underline{1.32}/0.36 & \textbf{1.10}/\textbf{0.36} & \underline{1.15}/\underline{0.36} \\
        \hline
        \multirow{2}{*}{\rotatebox[origin=c]{90}{ours}} & \textbf{RaNDT~SLAM} w/o LC & 1.51/0.49 & 1.53/0.50 & 1.55/0.50 & 1.49/0.51 & 1.31/0.46 & 1.51/0.47 & 2.03/0.50 & 1.92/0.65 & 1.60/0.52 \\
        ~ & \textbf{RaNDT~SLAM} w. LC & 1.52/0.50 & 1.54/0.49 & 1.40/0.46 & 1.45/0.49 & 1.19/0.44 & 1.44/0.46 & 1.72/0.52 & 1.72/0.60 & 1.50/0.50 \\
        \hline
        \hline
        ~ & ~ & \multicolumn{8}{c|}{\textbf{ATE (in m) on the Oxford dataset comparing SLAM methods.}} & ~\\
        \hline
        \multicolumn{2}{m{.2\textwidth}|}{RadarSLAM-Full \cite{Hong.2022}} & 9.59 & 11.18 & 5.84 & 21.21 & 7.74 & 13.78 & 7.14 & 6.01 & 10.31 \\
        \multicolumn{2}{m{.2\textwidth}|}{MAROAM \cite{Wang.25.10.2022}} & 13.76 & 6.95 & 8.36 & 10.34 & 10.96 & 12.42 & 12.51 & 7.71 & 10.38 \\
        \multicolumn{2}{m{.2\textwidth}|}{TBV Radar SLAM-dyn-cov \cite{Adolfsson.2023}} & \underline{4.22} & 4.30 & \textbf{3.37} & \underline{4.04} & 4.27 & \underline{3.38} & \underline{4.98} & \textbf{4.27} & \underline{4.10} \\
        \multicolumn{2}{m{.2\textwidth}|}{TBV Radar SLAM \cite{Adolfsson.2023}} & \textbf{4.07} & \underline{4.04} & \underline{3.58} & \textbf{3.79} & \textbf{3.83} & \textbf{3.14} & \textbf{4.39} & \underline{4.33} & \textbf{3.90} \\
        \hline
        \multicolumn{2}{m{.2\textwidth}|}{\textbf{RaNDT~SLAM} w. LC (ours)} & 5.40 & \textbf{4.00} & 5.89 & 6.45 & \underline{4.03} & 5.30 & 10.36 & 6.27 & 5.96 \\
        \hline
    \end{tabular}
    \label{tab:oxford_results}
\end{table*}

\begin{figure*}[t!]
    \def\picwidth{0.22\textwidth}
    \begin{center}
        \begin{subfigure}{\picwidth}
            \centering
            \includegraphics[width=\linewidth]{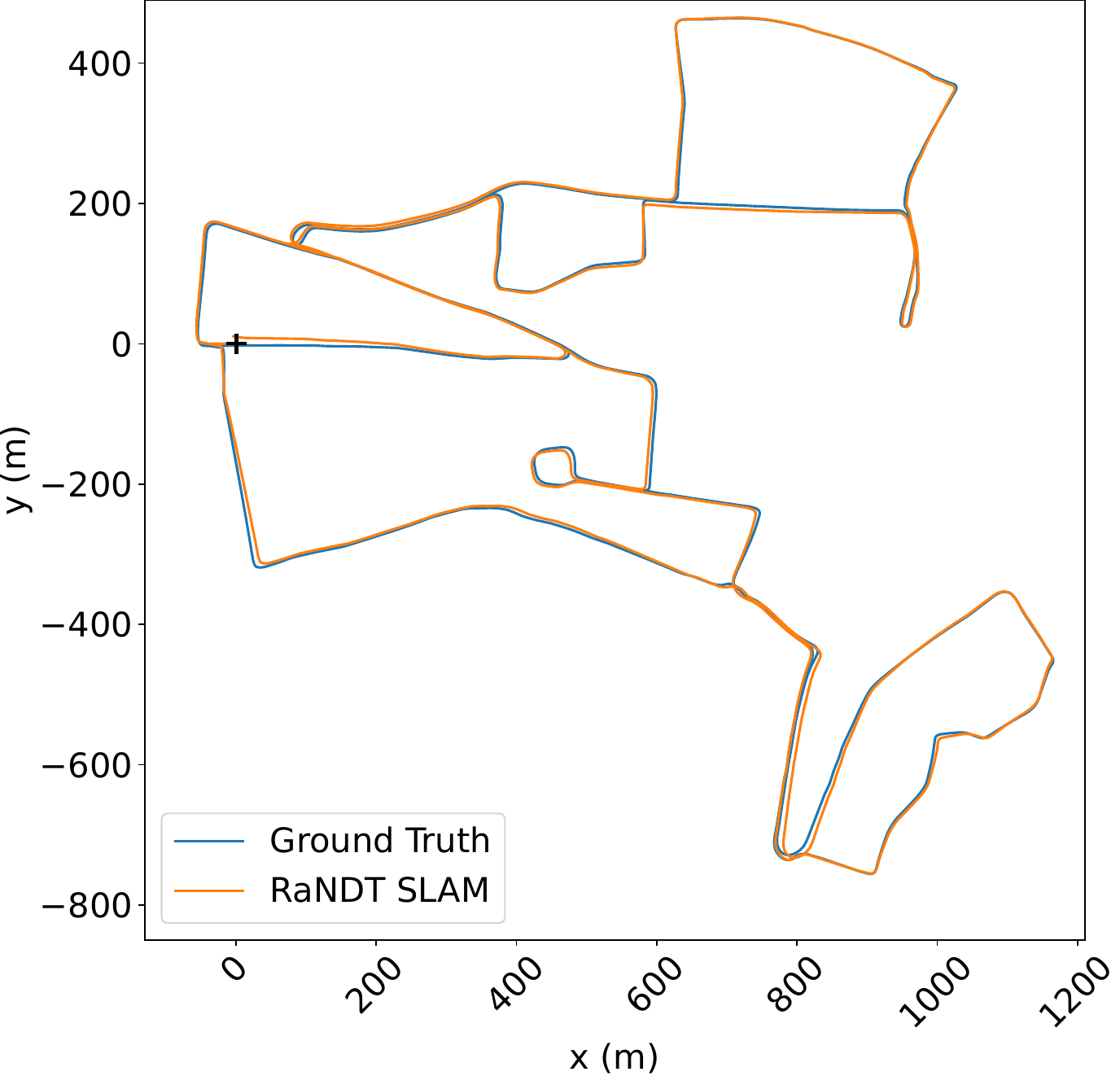}
            \caption{\texttt{10-12-32}}
        \end{subfigure}
        \begin{subfigure}{\picwidth}
            \centering
            \includegraphics[width=\linewidth]{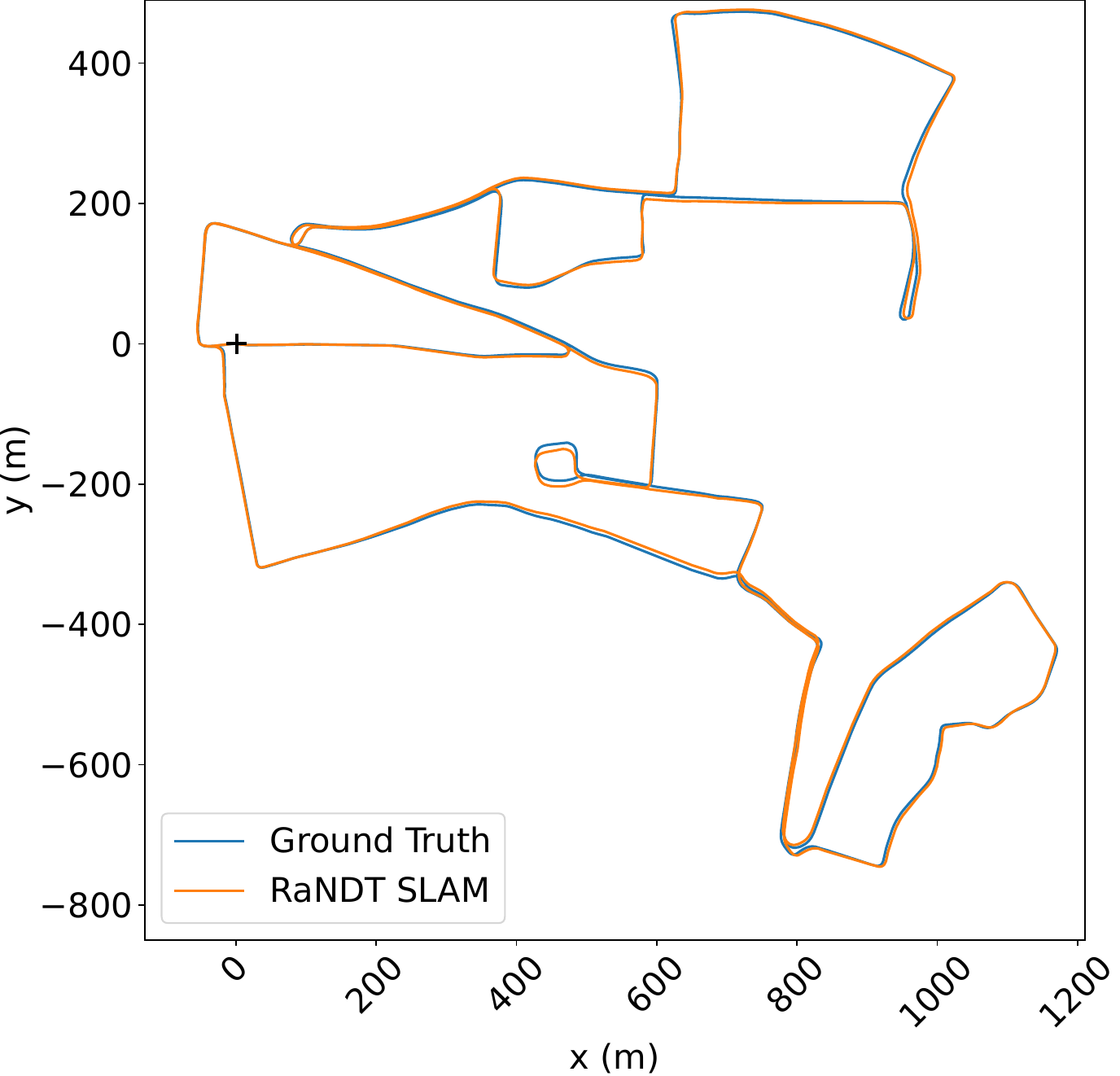}
            \caption{\texttt{16-13-09}}
        \end{subfigure}
        \begin{subfigure}{\picwidth}
            \centering
            \includegraphics[width=\linewidth]{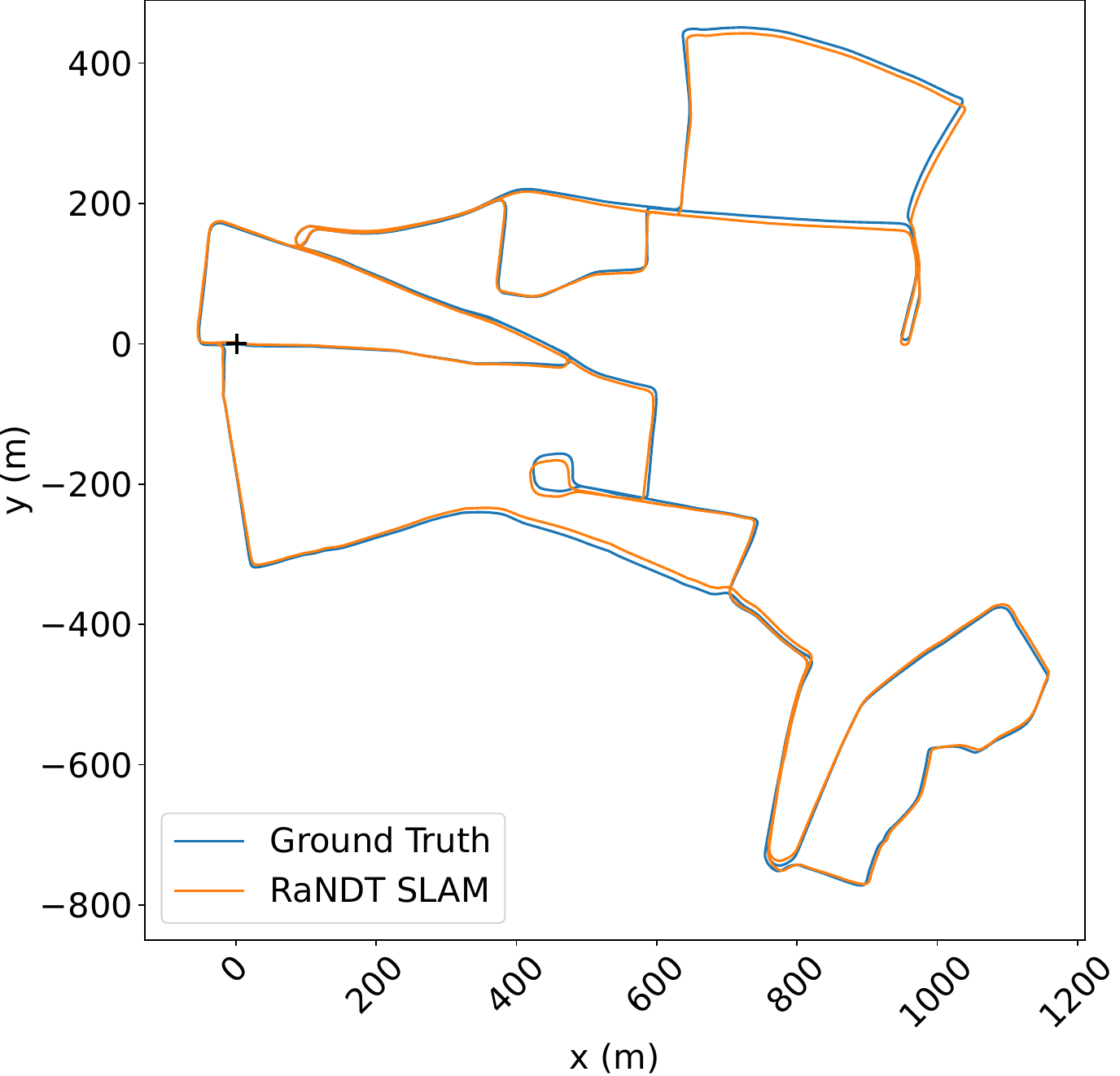}
            \caption{\texttt{17-13-26}}
        \end{subfigure}
        \begin{subfigure}{\picwidth}
            \centering
            \includegraphics[width=\linewidth]{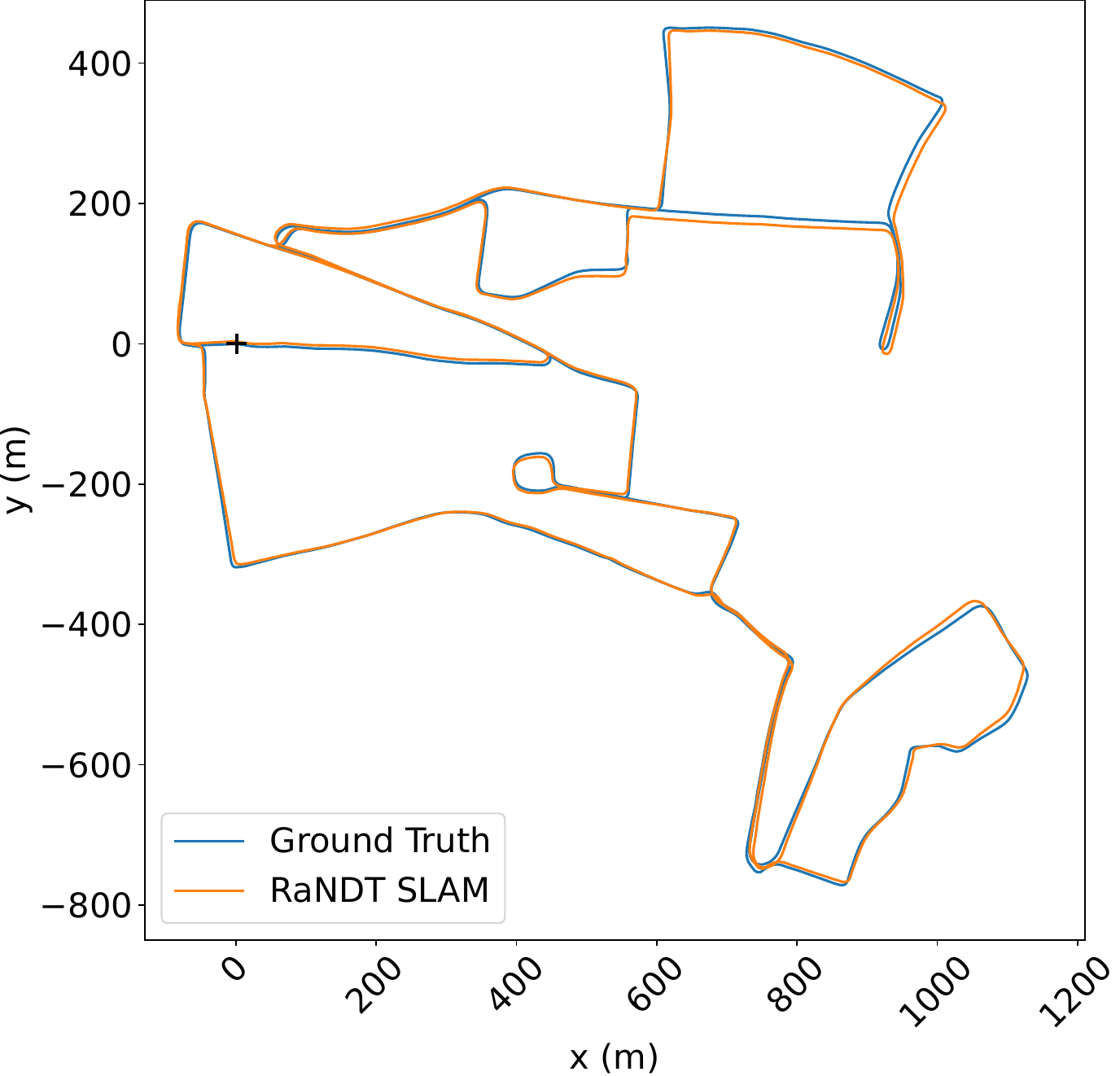}
            \caption{\texttt{18-14-46}}
        \end{subfigure}
    \end{center}
    \caption{Selected Oxford trajectories estimated by RaNDT SLAM compared to GT. Start positions are marked (+).}
\end{figure*}

\subsection{Computational Cost}
To assess the real-time capability, we evaluate the average duration spent by the most important subsystems (Table~\ref{tab:computational_costs}). We measure the graph optimization each time new loop closure constraints are added to the graph. No robust measure for the standard deviation can be given in the outdoor sequence, as only two loop closures are detected in the trajectory. The pre-processing and motion estimation steps are performed synchronous. Therefore, the time spent on a complete cycle of RaNDT~SLAM is significantly smaller than the sensor's measurement cycle of $200$~ms. This allows the online usage for autonomous operation and is promising for future radar sensors with higher scanning rate. The increased processing times in the indoor application are caused by the finer resolution of the NDT map. This leads to NDTs with more components, which makes the optimization more time consuming. Moreover, we observe that the indoor scenario takes approximately twice as long. This is because more distributions are generated, which makes the optimization harder. Note, that the cycle times are parameter-dependent and need to be considered as a trade-off between computational times, resolution and expected noise ratio.

\subsection{Evaluation with Oxford Radar RobotCar Dataset}
\label{ssec:oxford_eval}
To underline the applicability of RaNDT SLAM in different environments and with different sensor models, we test RaNDT~SLAM on the Oxford dataset~\cite{Barnes.2020b}, using the provided GT obtained via visual odometry, visual loop closures, and GPS constraints. This allows us to compare our results with state-of-the-art radar SLAM and radar odometry approaches, partially introduced in Section~\ref{sec:sota}. Table~\ref{tab:oxford_results} depicts the KITTI odometry errors and the ATEs. As the ATE largely favors algorithms using loop closure, we present only full SLAM systems and omit results obtained with radar odometry. The ATE achieved with RaNDT~SLAM scores second-best just behind TBV SLAM \cite{Adolfsson.2023}. In addition, the translation drift is able to keep up with other radar SLAM and radar odometry methods. Notably, we perform better than the other NDT-based methods regarding the translational drift~\cite{Hong.2022, Zhang.2023}. The rotational drift, however, is less competitive. We attribute this to the missing motion compensation in our framework, as motion distortion compromises the measurements especially during turns of the vehicle. We conclude, that RaNDT~SLAM generalizes well to different domains and sensor types, although being designed for smaller-scale and indoor environments. We expect the performance to increase further if techniques tailored for autonomous driving datasets such as motion compensation~\cite{Burnett.2020} or origin shifting in descriptor generation~\cite{Kim.2022} are implemented.

\section{CONCLUSION AND FUTURE WORK}
\label{sec:conclusion}
In this work, we proposed a novel radar SLAM framework, RaNDT~SLAM, suited for both indoor and outdoor environments. Within, we combined NDT registration, IMU measurements, and a motion model. We validated our approach using indoor and outdoor scenarios and reached high trajectory accuracy in both cases. Furthermore, we evaluated the computational effort and showed that we achieve real-time performance. This paves the way to robust, autonomous operation in small-scale, vision-denied environments. Further, we compared RaNDT~SLAM to the state of the art in large-scale radar SLAM and odometry using the Oxford Radar RobotCar dataset, which underlines its good performance even outside its designed scope. Code\footref{ft:2} and benchmark dataset~\cite{HM23} have been made publicly available, setting a new baseline for radar SLAM in rescue robotics environments. Possible extensions of our work may integrate more sensor data to improve the accuracy in case of low vision restriction or environments with switching conditions. The flexible estimator structure can easily be extended, e.g., using Doppler measurements. We aim to deploy RaNDT~SLAM with the German Rescue Robotics Center, to access its performance in application.

\addtolength{\textheight}{-0.0cm}   




\section*{ACKNOWLEDGMENT}
We like to thank indurad, particularly Matthias Rabel and Stephan Renner, for providing the iSDR-C sensors.


\bibliographystyle{IEEEtran}
\bibliography{IEEEabrv,References.bib}

\end{document}